\documentclass[utf8]{FrontiersinHarvard} % for articles in journals using the Harvard Referencing Style (Author-Date), for Frontiers Reference Styles by Journal: https://zendesk.frontiersin.org/hc/en-us/articles/360017860337-Frontiers-Reference-Styles-by-Journal
\usepackage{url,hyperref,lineno,microtype,subcaption,comment}
\usepackage[onehalfspacing]{setspace}

\def\keyFont{\fontsize{8}{11}\helveticabold }
\def\firstAuthorLast{Hanifi {et~al.}} %use et al only if is more than 1 author
\def\Authors{Shiva Hanifi\,$^{1,*}$, Elisa Maiettini\,$^{1,\dagger}$, Maria Lombardi\,$^{1,\dagger}$, and Lorenzo Natale$^{1}$}
\def\Address{$^{1}$Humanoid Sensing and Perception Group, Istituito Italiano di Tecnologia, Genoa, Italy.\\
$\dagger$These authors contributed equally to the work.}
\def\corrAuthor{Corresponding Author}

\def\corrEmail{shiva.hanifi@iit.it}

\begin{document}
\onecolumn
\firstpage{1}

\title[iCub Detecting Gazed Objects]{iCub Detecting Gazed Objects: A Pipeline Estimating Human Attention} 

\author[\firstAuthorLast ]{\Authors} %This field will be automatically populated
\address{} %This field will be automatically populated
\correspondance{} %This field will be automatically populated

\extraAuth{}% If there are more than 1 corresponding author, comment this line and uncomment the next one.
%\extraAuth{corresponding Author2 \\ Laboratory X2, Institute X2, Department X2, Organization X2, Street X2, City X2 , State XX2 (only USA, Canada and Australia), Zip Code2, X2 Country X2, email2@uni2.edu}

%TC:ignore
\maketitle

\begin{abstract}

%%% Leave the Abstract empty if your article does not require one, please see the Summary Table for full details.

%%% For full guidelines regarding your manuscript please refer to \href{http://www.frontiersin.org/about/AuthorGuidelines}{Author Guidelines}.

%%% As a primary goal, the abstract should render the general significance and conceptual advance of the work clearly accessible to a broad readership. References should not be cited in the abstract. Leave the Abstract empty if your article does not require one, please see \href{http://www.frontiersin.org/about/AuthorGuidelines#SummaryTable}{Summary Table} for details according to article type. 

\noindent
This research report explores the role of eye gaze in human-robot interactions and proposes a learning system for detecting objects gazed at by humans using solely visual feedback. The system leverages face detection, human attention prediction, and online object detection, and it allows the robot to perceive and interpret human gaze accurately, paving the way for establishing joint attention with human partners. Additionally, a novel dataset collected with the humanoid robot iCub is introduced, comprising over $22,000$ images from ten participants gazing at different annotated objects. This dataset serves as a benchmark for the field of human gaze estimation in table-top human-robot interaction (HRI) contexts. In this work, we use it to evaluate the performance of the proposed pipeline and examine the performance of each component. Furthermore, the developed system is deployed on the iCub, and a supplementary video showcases its functionality. The results demonstrate the potential of the proposed approach as a first step to enhance social awareness and responsiveness in social robotics, as well as improve assistance and support in collaborative scenarios, promoting efficient human-robot collaboration. 
%Data and code will be released upon acceptance.
% abstract is 183 words (max for a brief research report is 250 words)

\tiny
 \keyFont{ \section{Keywords:} attention, gaze estimation, learning architecture, humanoid robot, computer vision, human-robot scenario} %All article types: you may provide up to 8 keywords; at least 5 are mandatory.
\end{abstract}
%TC:endignore

%%%%%%%%%%%%%%%%%%%%%%%%%%%%%%%%%%%%%%%%%%%%%%%%%%%%%%%%%%%%%%%%%%%%%%%%%%%%%%%%
\section{Introduction}
\label{sec:introduction}

Any face-to-face interaction between two people is characterised by a continuous exchange of social signals, such as gaze, gestures, and facial expressions. Such non-verbal communication is possible because both interacting individuals are able to see each other, perceive and understand the social information enclosed in cues. In this study, we focus on one of the most crucial social cues, eye gaze. Eye gaze plays a pivotal role in many mechanisms of social cognition, e.g. joint attention, regulating and monitoring turn-taking, signaling attention and intention. Neuropsychological evidence highlighted the close relationship between gaze direction and attention, indicating that gaze functions are actively involved and influenced by spatial attention systems \cite{allison2000, pelphrey2003}. For example, it is more likely that the gaze is directed toward an object rather than toward empty space.

In this context, a robot's ability to determine what the human is looking at (e.g., an object) has numerous practical implications across various domains. In social robotics, it enhances a robot's social awareness and responsiveness, making interactions more natural and context-appropriate \cite{babel2021, holman2021}. This includes recognizing a person's preferences based on their gaze and improving collaboration in settings like industry or home by understanding human attention \cite{kurylo2019}.

This research report represents the initial milestone in our ongoing study that aims at interpreting the human intent, during a human-robot collaboration. In this work, we present a novel application for HRI that leverages computer vision techniques to enable robots to detect the object the human partner is gazing at. This application sets the baseline for forthcoming advancements in our research. Our proposed system combines an online object detection algorithm~\cite{ceola2020segm,Maiettini2019a} with gaze tracking technologies, providing the robot with online information about the objects that capture the human's attention. This integration offers a valuable cognitive capability for the robot, empowering it to accurately perceive and interpret the human gaze within its environment. That can be the first step for making the robot able to establish conscious joint attention with the human partner \cite{chevalier2020}.

The main contributions are as follows:

\begin{itemize}
    \item We propose a pipeline to detect the target of human attention during an interaction with a robot. This leverages face detection, human attention prediction, and online object detection to detect the object the human focuses on.
    \item We present the \textit{ObjectAttention} dataset collected with the humanoid iCub~\cite{icub} where $10$ participants gaze at different objects placed randomly on a table in front of the robot, totaling an amount of over $22K$ images. This dataset serves as a benchmark for application performance evaluation.
    \item We perform an experimental analysis of the proposed pipeline to evaluate its effectiveness in the considered HRI setting. To do that we use the collected dataset and we study the performance of the components of the system.  
    \item Finally, we deploy the system on the iCub robot. The video submitted as Supplementary Material shows its functioning. 
\end{itemize}
%%%%%%%%%%%%%%%%%%%%%%%%%%%%%%%%%%%%%%%%%%%%%%%%%%%%%%%%%%%%%%%%%%%%%%%%%%%%%%%%
\section{RELATED WORK}
\label{sec:related_work}

The problem of endowing robots with the ability to understand human behavior and specifically the social cue of the gaze has been largely studied in the literature. It is mainly addressed following two different strategies: 1) gaze estimation (i.e. estimating the gaze vector or mutual gaze events) and 2) gaze attention prediction (i.e. understanding where the human is visually attending in terms of saliency map).

Following the \textit{gaze estimation strategy}, a learning architecture to detect events of mutual gaze was proposed in \cite{lombardi2022-mutualgaze}. This study emphasized the importance of mutual gaze as a crucial social cue in face-to-face interactions since it can be a signal of the readiness of the interacting partner. Several works focused, instead, on the estimation of the human gaze as a 2D or 3D vector. In \cite{krafka2016} a CNN was used to extract features from the RGB image (face and both eyes images) and produce as output the $(x,y)$ coordinate of the gaze prediction. The use of the CNN architecture to estimate the 2D gaze vector is also proposed in \cite{athavale2022}, but with the novelty of extracting features from only one eye, that is especially useful in real-world conditions where the human face can be partially obscured. An example of predicting a 3D gaze vector can be found in \cite{cheng2020}. The authors proposed a combination of a regression network (FAR-net) and an evaluation network (E-Net) able to exploit the asymmetry and the difference between left and right eye.

Much effort has been spent in building \textit{attention architectures} that, for example, enable joint attention between a human and the iCub robot with the aim to improve the performance of visual learning methods \cite{Lombardi2022}. Yet, in \cite{sumer2020} the authors proposed a method, called Attention Flow, to learn joint attention in an end-to-end fashion by employing saliency-augmented attention maps and two convolutional attention mechanisms that choose important cues and improve joint attention localization. %Moreover, in \cite{fan2018}, a spatial-temporal neural network (LSTM network) was proposed to detect shared attention intervals in third-person social videos and predict shared attention locations in frames.

The \textit{gaze following} problem was addressed, for example, in \cite{recasens2017}. %and \cite{chong2018}. 
A CNN architecture was proposed taking as input the RGB frame (with the person's head location and the eye coordinates in that frame) and a set of neighboring frames from the same video and identifying which of the neighboring frames, if any, contain the object being looked at and the coordinates of the person’s gaze. 
%The latter proposed a two-pathway solution: the gaze direction pathway takes as input the head image and position and generates multi-scale gaze direction fields by using ResNet-50 as feature extractor; in the  heatmap pathway the framework tries to infer the gaze point taking into account the gaze direction and the context information of the objects along the gaze direction.

Despite gaze estimation and human attention have been largely studied, few work exist integrating human attention estimation with target object prediction. Among these few, authors in~\cite{saran2018} proposed an approach to predict human referential gaze having both the person and object of attention visible in the image. The proposed network contains two pathways: one estimating the head direction and another for salient objects in the scene. Such a network was used as backbone in \cite{vtd} and extended to address out-of-frame gaze targets by simultaneously learning gaze angle and saliency. Moreover, differently from~\cite{saran2018}, an LSTM-based spatio-temporal model is used to leverage the temporal coherence of the frames in videos to improve gaze direction estimation. However, in~\cite{vtd}, only the human gaze direction is predicted, while the information about the target object is not provided. 

%In this paper, we aim at improving the system and the performance stated in \cite{vtd} and making also the pipeline run online on the humanoid robot iCub. This deployment makes the robot capable of inferring where the human partner’s attention is targeted while interacting with him/her.

In this research report, we exploit the LSTM-based spatio-temporal model presented in~\cite{vtd} and we adapt it to the considered HRI setting by fine-tuning it on the proposed ObjectAttention dataset. Moreover, we use it in conjunction with a human pose estimation and a face detector to make the pipeline run online on the iCub robot. Finally, by integrating an online object detection method, we allow the system to predict the class label and location of the gaze target object. Note that, differently from~\cite{saran2018}, by using~\cite{Maiettini2019b} for object detection, the entire system can be easily and quickly (just a few seconds) adapted to detect novel target objects.
All the mentioned improvements result in an online robotic application that makes the robot capable of inferring where the human partner’s attention is targeted while interacting with them.
%%%%%%%%%%%%%%%%%%%%%%%%%%%%%%%%%%%%%%%%%%%%%%%%%%%%%%%%%%%%%%%%%%%%%%%%%%%%%%%%
\section{METHODS}
\label{sec:methods}

\begin{figure}[h!]
    \centering   \includegraphics[width=0.9\textwidth]{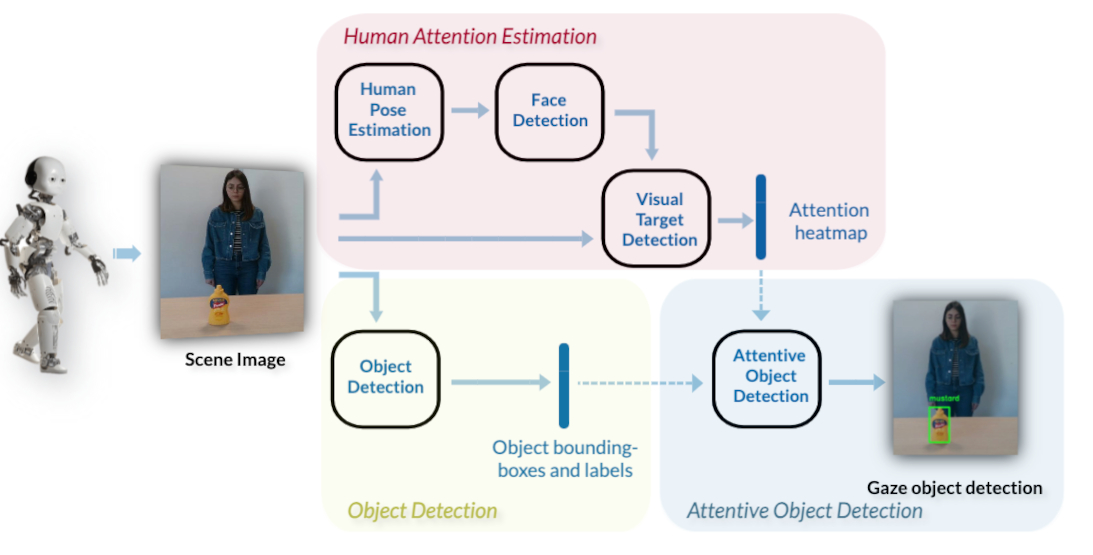}
    \caption {Pipeline of the presented architecture. The \textit{Human Attention Estimation} pathway produces a heatmap of the gaze target. This heatmap as well as the bouding boxes and labels from the \textit{Object Detection} module are then used by the \textit{Attentive Object Detection} module to predict the specific object that is visually attended by the human.}
    \label{fig:pipeline}
\end{figure}

The proposed pipeline is made out of three pathways (see Fig.~\ref{fig:pipeline}): the \textit{Human Attention Estimation} pathway that aims at detecting the attention target of the human, the \textit{Object Detection} pathway that recognizes and localizes the objects in the scene, and finally the \textit{Attentive Object Detection} pathway that provides the gazed object from the human.

\subsection[Human Attention Estimation]{Human Attention Estimation}\label{subsec:Online Visual Target Detection}
The \textit{Human Attention Estimation} pathway is characterised mainly by three distinct modules: 1) Human Pose Estimation, 2) Face Detection, and 3) Visual Target Detection. Having the RGB image as input, the final output of this pathway is the real-time prediction of the human attention target, provided in the form of heatmap. %Each module is described below and depicted in  Figure \ref{fig:pipeline}.\\

\noindent\textbf{Human Pose Estimation}. We rely on the OpenPose architecture, proposed in~\cite{openpose2019}. Briefly, OpenPose is a system for multi-human pose estimation, that receives as input RGB frames and predicts the location in pixel $(x,y)$ of $135$ anatomical key-points of each person in the image, associating also a confidence level $k$ to each prediction. %In our pipeline, we use this approach to detect the human in front of the robot.\\ %($25$ and $70$ key-points for the body pose and for the face, respectively)

\noindent\textbf{Face Detection}. We rely on the face recognition presented in \cite{Lombardi2022} to detect and extract the human face from the image. Specifically, the face key-points extracted by the \textit{Human Pose Estimation} are used as input while the output is the bounding box of the head of the person in front of the robot. Note that, in~\cite{vtd}, they assume that the information of the face location is available, reading it from a txt file. That is a strong limitation in applying the method in online robotic applications, preventing it from being used on real robots. In this work, we provide the online input to the \textit{Visual Target Detection module} by using the \textit{Human Pose Estimation} together with the \textit{Face Detection}, enabling the pipeline to operate on the actual robot.\\

\noindent\textbf{Visual Target Detection}.This module takes as input the RGB image from the robot camera and the human face bounding box extracted by the \textit{Face Detection} module. It provides as output the heatmap representing the image area that more likely contains the target of human attention. Specifically, this is a matrix of the same dimensions of the image, where each cell corresponds to a pixel in the image. The value of each cell ranges from 0 to 1 (respectively, the lowest and the highest probability to be --or to be close to-- the target of human attention). For this module, we rely on the network presented in~\cite{vtd}, which is composed of three main parts. The first one is \textit{Head Conditioning Branch} which uses the head bounding box, encoded into a convolutional feature map (head feature map) together with the information of the location of the human's head in the image to predict a first attention map. The second part is the \textit{Main Scene Branch}, which multiplies the convolutional feature map of the entire image with the attention map and concatenates the result with the previously computed head feature map. The final tensor represents the input for the third and last part, namely, the \textit{Recurrent Attention Prediction Branch}. This, firstly encodes the tensor in order to be used as input for a convolutional Long Short-Term Memory network. The final attention heatmap is then created by upsampling the output of the latter using a decoder. 
In this work, we fine-tune the weights of the network by using our dataset, and the resulting model is used for the developed application and for the experimental analysis. %(details are reported in Sec.~\ref{sec:dataset} and~\ref{subsec:training}).

\subsection[Object detection]{Object detection}\label{subsec:Object detection}
The \textit{Object Detection} pathway is mainly characterized by one module that takes the RGB images from the camera of the robot as input and gives as output the bounding boxes of all the objects of interest present in the scene. For this task, we rely on the Online object detection approach, presented in~\cite{ceola2020segm,Maiettini2019a}. This system is based on Mask R-CNN architecture conveniently adapted so that it can be re-trained online in a few seconds, allowing for a fast adaptation without a performance loss. We train the Online object detection with data acquired using the pipeline described in~\cite{maiettini2017}. %(details are in Sec.~\ref{subsec:training}).

\subsection[Attentive Object detection]{Attentive object detection}\label{subsec:Attentive object detection}
The third pathway combines the extracted information from human attention with the objects in the scene to detect the object that is the target of the human gaze. It takes as input the RGB image, the heatmap from the \textit{Visual Target Detection Module}, and all the bounding boxes and labels predicted by the \textit{Object Detection}. The output of this module is the attended object bounding box and label.

Firstly, the heatmap is processed to identify the contour of the area with the highest values which corresponds to the image area where the human is focusing their gaze (the hottest part of the heatmap). Then, we compute the center of the obtained area and the surrounding bounding box. We use this information to select the object that, most likely, is the focus of human attention. Precisely, we choose the object that either presents the smallest Intersection over Union (IoU) with the bounding box of the hottest part of the heatmap or, if this latter does not intersect any object bounding box, we select the one whose center is the closest to the center of the hottest part.
%%%%%%%%%%%%%%%%%%%%%%%%%%%%%%%%%%%%%%%%%%%%%%%%%%%%%%%%%%%%%%%%%%%%%%%%%%%%%%%%
\section{DATASET}
\label{sec:dataset}
%\input{Sections/Dataset}

%In this section we describe our novel ObjectAttention dataset that is collected specifically for the purpose of this research. 
A major contribution of this work is the \textit{ObjectAttention} dataset. It mainly depicts human-robot interactions in a table-top scenario where the human gazes at different objects and the robot understands the gaze direction and the target object. %The dataset was used to train, evaluate, and validate our architecture. %and will be accessible for reproducibility. %In the following sections, we detail the data collection and annotation procedures.

\subsection[Data collection sessions]{Data collection}\label{subsec:Data collection sessions}
We recruited a total of $10$ participants ($4$ females, $6$ males) with normal or corrected vision. The data collection was conducted with the iCub robot~\cite{icub}, and all participants provided written informed consent.
To collect the dataset, iCub was positioned, with a RealSense 415 camera\footnote{https://www.intelrealsense.com/depth-camera-d415/}mounted on its head, on one side of a table. On the table, we placed up to five objects from the YCB dataset~\cite{ycb}, in various arrangements and different for all the participants. For different sessions, the participants were instructed to stand on the other side of the table facing the robot and looking at the requested object in a natural and spontaneous manner. The frames were recorded using the RealSense 415 camera and the YARP middleware~\cite{yarp}.

We collected data in $5$ different sessions for each participant, starting with one object in the scene and gradually increasing the number of objects up to five. For each session, we performed two trials, with two different settings obtained keeping the same number of objects but changing the object types and their arrangements on the table. For each session and for each trial, each object is gazed at for a 5-second period by the participant and the gaze target ground truth is annotated considering the instructions given to the participant. We collected a $5$-seconds video for each different object.

The resulting dataset consists of $250$ videos (for a total of $22,732$ frames), depicting $10$ participants in $2$ different trials for each of the $5$ sessions, gazing at the different objects. Additionally, for at least one trial in each of the sessions, we placed a distracting object (i.e., the \textit{pringles} object) on the table, at which the participant was not asked to gaze. Example frames for three participants for different sessions are reported in the Supplementary Materials. %Fig.~\ref{fig:dataset_example_frames}.

\begin{figure}[h!]
    \centering
    \includegraphics[width=\textwidth]{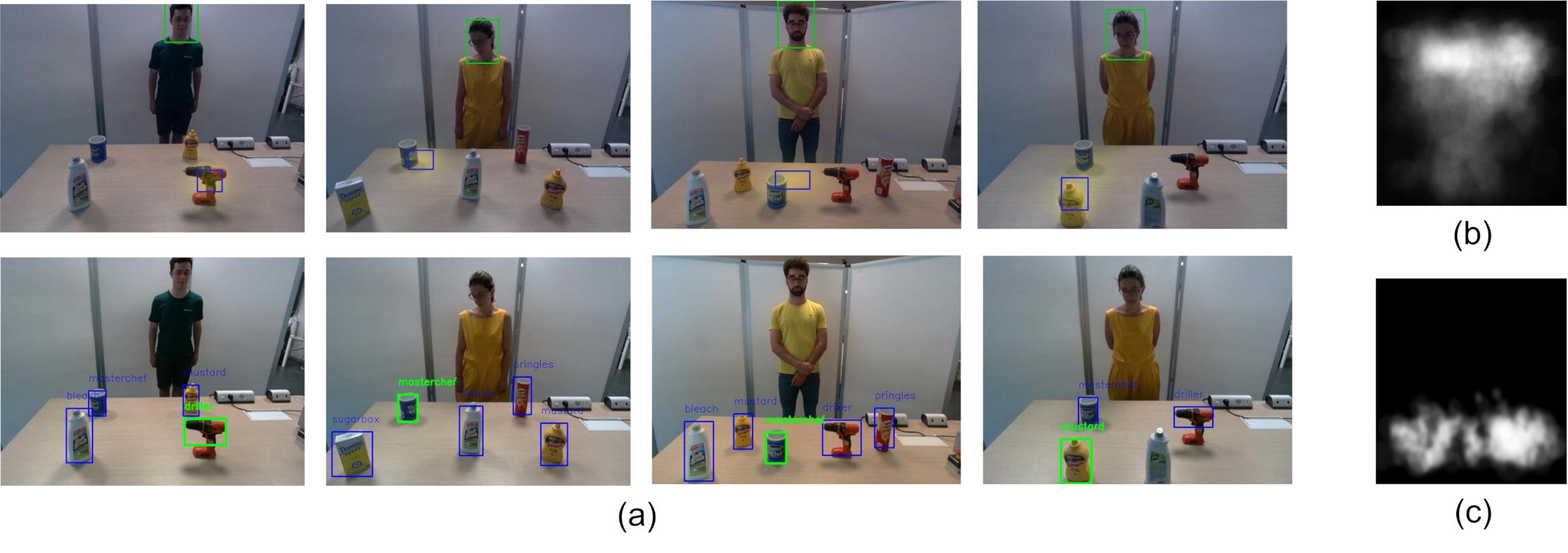}
    \caption{(a) Selection of sample output frames of the proposed pipeline. The first row depicts the scene image as well as the head bounding box of the participant detected by the \textit{Face Detection} module, the attention heatmap of the participant, and the bounding box of the hottest area of the heatmap. While the second row depicts the related gaze target selections for the frames of the first row. (b) Gaze target location density for the dataset in~\cite{vtd} and (c) the density map of the gaze targets for our dataset.}
    \label{fig:model_output}
\end{figure}

Finally, a motivation that have driven us to collect a new dataset is the fact that the dataset of~\cite{vtd} contains more conditions in which the gaze was directed toward the upper part of the map (and so not suitable for a table-top). Our dataset, in contrast, was collected for scenarios where the human and robot are looking at objects placed on a table. Fig.~\ref{fig:model_output} depicts the density map of the gaze targets for the dataset in~\cite{vtd} (b) and the one we collected (c). To improve performance in the considered setting, the method proposed in~\cite{vtd} was fined-tuned using our dataset.

\subsection[Data annotation]{Data annotation}\label{subsec:Data annotation}
For each setting, the bounding box of the participant's head and the target object are required. 
The participants' head bounding box was extracted using the key-points estimated by \textit{Openpose}~\cite{openpose} and manually refined to be considered as ground truth. Furthermore, we manually annotated the bounding boxes and classes for all the objects on the table, highlighting the one that is the target of the participant's attention. The gaze target point was chosen as the center of the gazed object. The bounding boxes labeling was done using the \textit{LabelImg}\footnote{\url{https://github.com/tzutalin/labelImg}} framework.
%%%%%%%%%%%%%%%%%%%%%%%%%%%%%%%%%%%%%%%%%%%%%%%%%%%%%%%%%%%%%%%%%%%%%%%%%%%%%%%%
\section{EXPERIMENTS}
\label{sec:experiments}

\subsection{Model training}
\label{subsec:training}
For the experiments reported in this research report, two different modules were re-trained to better fit the considered conditions, namely, the \textit{Object Detection} and the \textit{Visual Target Detection}.\\

\noindent\textbf{Object Detection training}. We trained the Online object detection with data acquired using the pipeline described in~\cite{maiettini2017}. Specifically, a human teacher showed the objects of interest to the robot, one at a time, holding them in their hand and moving them in front of the robot for around $30$ seconds. The information from the robot's depth sensors was used to localize the object and follow it with the robot’s gaze. The latter can be segmented and the corresponding bounding box was automatically assigned and gathered as ground truth together with the object's label, provided verbally. After each object demonstration, the collected data was used to update the current object detection model, optimizing its weights in a few seconds.\\

\noindent\textbf{Visual Target Detection fine-tuning}. To fine-tune the Visual Target Detection module, we randomly split the ObjectAttention dataset by participants, considering approximately $70\%$ of the dataset (data from $7$ participants) as trainset, and the remaining $30\%$ as testset, ensuring no overlap of data between the train and test splits.
We fine-tuned the spatio-temporal model of the \textit{Visual Target Detection} on the trainset performing a warm training re-start with the pre-trained weights provided by the authors~\cite{vtd}, and empirically choosing the hyper-parameters as follows: \textit{learning rate} $ = 5e^{-5}$, \textit{batch size}$ = 4$, \textit{chunk size}$ = 3$, \textit{number of epochs}$ = 10$. 
To ensure the statistical relevance of the presented experiments, we repeated the training and evaluation of the model for $3$ times with $3$ different splits of the dataset. 
%We report in Tab.~\ref{tab:vtd_evaluation} performance in terms of mean and standard deviation for the considered metrics over all the repetitions.

\subsection{Experimental Setup}
\label{subsec:exp_setup}
The performance of the \textit{Visual Target Detection} is evaluated in terms of the \textit{Area Under the Curve} (AUC) and \textit{Distance} metrics. For the first metric, each cell in the spatially-discretized image is classified as either the gaze target or not. The ground truth comes from thresholding a Gaussian confidence mask centered at the human annotator’s target location. The final heatmap provides the prediction confidence score which is evaluated at different thresholds in the ROC curve. The AUC of this ROC curve is considered. The \textit{Distance} metric, instead, is defined as the $\L_2$ distance between the annotated target location and the prediction given by the pixel of the maximum value in the heatmap, with image width and height normalized to 1. The performance for the entire pipeline is measured in terms of \textit{Accuracy} of the detected gazed objects. For each image, the bounding box of the predicted gazed object is compared with the ground truth: if the gazed object is correctly identified, the prediction is counted as true positive, otherwise, it is predicted as false negative. 

To run the experiments, the \textit{Human Pose Estimation} module was run on an Alienware workstation equipped with an NVIDIA GTX 1080TI while the remaining modules were executed on another Alienware workstation with graphics NVIDIA GeForce RTX 3080 additionally equipped with an external NVIDIA GTX 2080TI.

\subsection{Visual Target Detection fine-tuning}
\label{subsec:vtd_finetuning}

We first analyze the impact of fine-tuning the \textit{Visual Target Detection} module on our dataset. In Tab.~\ref{tab:vtd_evaluation}, we report the performance comparison of the proposed model (row \textbf{Fine-tuning}) with the model presented in~\cite{vtd} (row \textbf{Pre-trained model}), in terms of mean and standard deviation  over the three dataset splits mentioned above. As can be seen, fine-tuning the model on the proposed ObjectAttention dataset allows us to get better performance. Specifically, the predicted hottest point in the heatmap is closer to the true gazed point of $\sim$0.04. Note that this is a relevant difference since the Distance metric is computed on an image with width and height normalized to 1. This result is also supported by the improvement of the \textit{AUC} metric of $5\%$. %Getting a precise heatmap is crucial in order to get a better selection of the bounding box of the gazed object. 
This improvement is due to the fact that the model is fine-tuned using a dataset that more closely represents the target scenario (table-top). Nevertheless, since the fine-tuned network has been initialised with the weights presented in~\cite{vtd}, the final model results to be still able to predict gaze direction different from those considered in the proposed dataset. For a qualitatively performance see the video attached as Supplementary Material.

\subsection{Accuracy evaluation}
\label{subsec:pipeline}
We study the performance of the overall pipeline on the proposed ObjectAttention dataset. Specifically, for the \textit{Visual Target Detection} module, we choose one of the models trained on the three different train/test splits and we use it in our pipeline. To get quantitative results, we consider as testset the same used to previously evaluate the fine-tuned model and as ground truth the bounding boxes and labels of the objects on the table and of the target gazed object.
%(i.e., the three participants that have not been used for the training)
%However, differently from the previous section, since we need to analyze the performance of the entire pipeline of detecting the target gazed object, we will consider, 

Firstly, we analyze the overall accuracy of the pipeline in detecting the gazed target object by the three different participants in the testset. Specifically, in our experiments the pipeline succeeds in detecting the right target object for the $79.5\%$ of the times. Note that, this accuracy number represents the performance of the integration of the \textit{Visual Target Detection}, the \textit{Object Detection}, and the \textit{Attentive Object Detection}.
% Firstly, we report in Tab.~\ref{tab:quantitative_model_evaluation}, the overall accuracy of the pipeline (second column in Tab.~\ref{tab:quantitative_model_evaluation}) and different performance for all the participants in the test set (third, fourth and fifth columns in Tab.~\ref{tab:quantitative_model_evaluation}). As it can be observed, the overall accuracy of the system in detecting the gazed target object is 79.5\% with no significant differences between the different participants.

Fig.~\ref{fig:model_output} illustrates a selection of sample frames from the output, including attention heatmaps and bounding boxes, highlighting the head of the participant (detected by the \textit{Face Detection} module) and the hottest areas of the heatmap in the frames of the top row, while the final gaze object bounding box and label are presented in the bottom row frames. The performance are qualitatively shown also in the video in the Supplementary Materials.

\subsection{Performance analysis}
\label{subsec:analysis}
%To further study the proposed pipeline, in the next paragraphs we present the analysis of the performance of the system for different objects, and sessions showing also the effect of the distractor object in the scene.\\

\begin{figure}[h!]
    \centering
    \includegraphics[width=\textwidth]{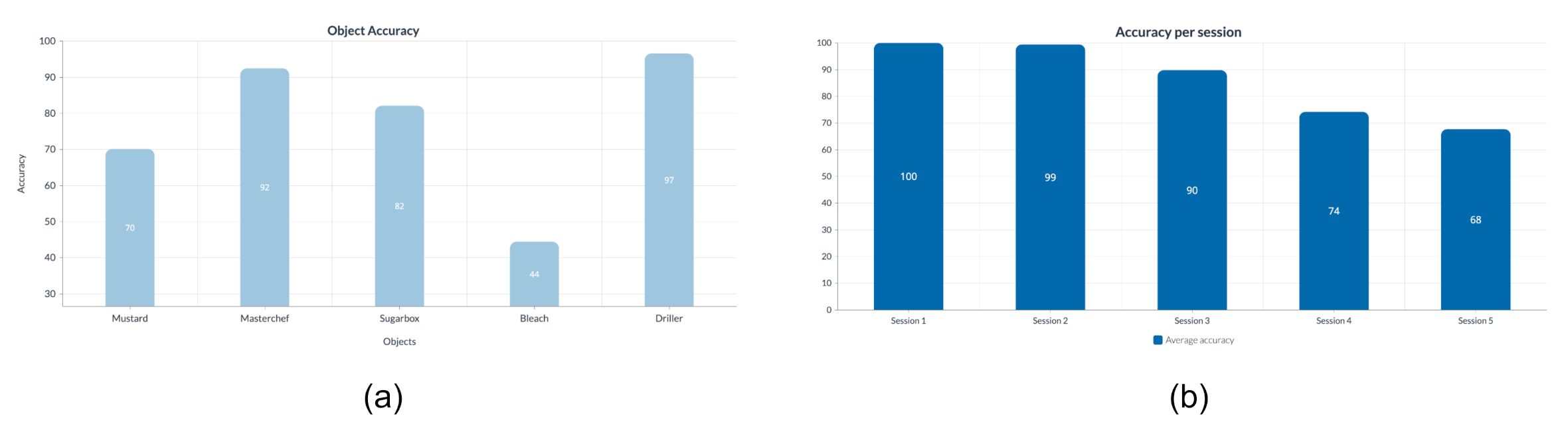}
    \caption{Accuracy analysis: (a) on each of the objects involved in the experiments, (b) on each of the sessions for all the participants. }
    \label{fig:accuracy}
\end{figure}

\noindent\textbf{Per object performance}. In Fig.~\ref{fig:accuracy} (b), we report the level of accuracy achieved when the different objects are considered as targets. As it can be observed, the system has a high performance for almost all the objects except for the object class \textit{Bleach}. This discrepancy is primarily attributed to difficulties in object detection, as sometimes the system fails to accurately locate the \textit{Bleach} object. This issue may stem from differences in the training conditions of the detector compared to the testing conditions, resulting in a domain shift. 
Prior research has suggested that addressing this problem can be achieved through methods such as integrating a robot's autonomous exploration of the new domain and employing weakly-supervised learning techniques ~\cite{Maiettini2020_thesis,Maiettini2019b,Maiettini2020}.\\

\noindent\textbf{Per session performance}. Fig.~\ref{fig:accuracy} (a) depicts the accuracy levels of the overall pipeline on various sessions. The performance of the system slightly decreases for higher numbers of session. This is reasonable since in those cases the number of objects in the scene increases thus the table becomes more cluttered. However, the accuracy level is still acceptable (around 70\%) even with the most cluttered scenes, showing that this is not a limitation of the proposed system.\\

\noindent\textbf{Distractors}. Finally, we report on the impact of the distracting objects on the system performance. While all objects on the table can be considered distracting when a person is focused on one of them, to analyze their impact, we select one sample object (i.e., the \textit{pringles}) as distractor in our dataset.  The \textit{Object Detection} module has been trained to detect it, even though the participants were not asked to gaze at it. We aim at verifying if the presence of the distractor object poses challenges in correctly identifying the target object. The results indicate that only in $\sim$3\% of the frames where the distracting object is present, it leads to errors in predictions, demonstrating that this is not a limitation of the proposed system.

%%%%%%%%%%%%%%%%%%%%%%%%%%%%%%%%%%%%%%%%%%%%%%%%%%%%%%%%%%%%%%%%%%%%%%%%%%%%%%%%
\section{CONCLUSIONS}
\label{sec:conclusions}

We presented a learning system for detecting human attention in relation to the objects in the scene. Our method combined an online object detection algorithm with a network for gaze estimation conditioned on the pose of the human, estimated using a pose estimation method. We demonstrated its effectiveness through an extensive experimental analysis using the iCub robot.
Our results indicated that the integration of face detection, human attention prediction, and online object detection in our pipeline enables the robot to perceive and interpret human gaze within its environment. Such an achievement is promising in enhancing the robot's social awareness and responsiveness, allowing for more natural interactions in social robotics.
The presented pipeline and dataset pose the baseline of an ongoing work aiming at endowing iCub with different social cues in a multimodal architecture, to improve the robot's skills in collaborative tasks. Future work will explore the use of the language signal to improve and extend the gaze target predictions.
%%%%%%%%%%%%%%%%%%%%%%%%%%%%%%%%%%%%%%%%%%%%%%%%%%%%%%%%%%%%%%%%%%%%%%%%%%%%%%%%
\section*{Data Availability Statement}
The code and the collected dataset will be released upon acceptance.

\section*{Ethics Statement}
The participants provided their written informed consent to participate in this study.

\section*{Author Contributions}
EM, ML and LN conceived the main idea of the study. EM and ML conceived and designed the learning architecture. SH, EM and ML collected the data. SH and EM implemented the learning system. SH performed the experiments and analysed the data. SH, EM and ML deployed the algorithm on the robot. SH, EM, ML and LN discussed the results. SH, EM and ML wrote the manuscript. All authors revised the manuscript.

\section*{Funding}
This work received funding from the Italian National Institute for Insurance against Accidents at Work (INAIL) ergoCub Project, and the project Fit for Medical Robotics (Fit4MedRob) - PNRR MUR Cod. PNC0000007 - CUP: B53C22006960001.

\section*{Conflict of Interest Statement}
The authors declare that the research was conducted in the absence of any commercial or financial relationships that could be construed as a potential conflict of interest.

%\section*{Supplementary Material}
%%%%%%%%%%%%%%%%%%%%%%%%%%%%%%%%%%%%%%%%%%%%%%%%%%%%%%%%%%%%%%%%%%%%%%%%%%%%%%%%

\bibliographystyle{Frontiers-Harvard}
%\bibliography{bibliography_elisa}  % .bib
\bibliography{bibliography} 

%%%%%%%%%%%%%%%%%%%%%%%%%%%%%%%%%%%%%%%%%%%%%%%%%%%%%%%%%%%%%%%%%%%%%%%%%%%%%%%%

\section*{TABLE CAPTIONS}

\begin{table}[h!]
    \vspace{0.25cm}
    \centering
    \label{tab:vtd_evaluation}
    \begin{tabular}{|c|c|c|}
      \hline
       \textbf{Method}&\textbf{ AUC (\%) $\uparrow$} & \textbf{$\L_2$ distance $\downarrow$} \\
      \hline
      \hline
      \textbf{Pre-trained model} & 87.5 $\pm$ 0.9 & 0.131 $\pm$ 0.014 \\
      \hline
      \textbf{Fine-tuning} & 92.5 $\pm$ 1.9 & 0.089 $\pm$  0.014\\
      \hline
    \end{tabular}
    \caption{Quantitative evaluation of the \textit{Visual Target Detection} model on the presented \textit{ObjectAttention} dataset.}
     \label{tab:vtd_evaluation}
\end{table}

\section*{FIGURE CAPTIONS}

% Accuracy evaluation per session and per object

%%%%%%%%%%%%%%%%%%%%%%%%%%%%%%%%%%%%%%%%%%%%%%%%%%%%%%%%%%%%%%%%%%%%%%%%%%%%%%%%

\end{document}